\documentclass{article}

\PassOptionsToPackage{numbers, compress}{natbib}

\usepackage[final]{nips_2018}

\usepackage[utf8]{inputenc} 
\usepackage[T1]{fontenc}    
\usepackage{hyperref}       
\usepackage{url}            
\usepackage{booktabs}       
\usepackage{amsfonts}       
\usepackage{nicefrac}       
\usepackage{microtype}      
\usepackage{graphicx}
\usepackage{amsmath}
\usepackage[linesnumbered,ruled, noend]{algorithm2e}
\usepackage{wrapfig}
\usepackage{bbm}

\title{Alternating Loss Correction for Preterm-Birth Prediction from EHR Data with Noisy Labels}

\author{
  Sabri Boughorbel \\
  Systems Biology\\
  Sidra Medicine\\
  Doha, P.O. Box 26999, Qatar \\
  \texttt{sboughorbel@sidra.org} \\
  \And
  Fethi Jarray \\
  Higher Institute of Computer  \\
  Science, Medenine, Tunisia \\
  \texttt{fjarray@gmail.com} \\
  \And
  Neethu Venugopal \\
  Systems Biology\\
  Sidra Medicine\\
  Doha, P.O. Box 26999, Qatar \\
  \texttt{nvenugopal@sidra.org} \\
  \And
  Haithum Elhadi\\
  Medical Informatics\\
  Sidra Medicine\\
  Doha, P.O. Box 26999, Qatar \\
  \texttt{helhadi@sidra.org} \\
}

\begin{document}

\maketitle
\vspace{-0.5cm}
\begin{abstract}
  In this paper we are interested in the prediction of preterm birth based on diagnosis codes from longitudinal EHR. We formulate the prediction problem as a supervised classification with noisy labels. Our base classifier is a Recurrent Neural Network with an attention mechanism. We assume the availability of a data subset with both noisy and clean labels. For the cohort definition,  most of the diagnosis codes on mothers' records related to pregnancy are ambiguous for the definition of full-term and preterm classes. On the other hand, diagnosis codes on babies' records provide fine-grained information on prematurity. Due to data de-identification, the links between mothers and babies are not available. We developed a heuristic based on admission and discharge times to match babies to their mothers and hence enrich mothers' records with additional information on delivery status. The obtained additional dataset from the matching heuristic has noisy labels and was used to leverage the training of the deep learning model. We propose an Alternating Loss Correction (ALC) method to train deep models with both clean and noisy labels. First, the label corruption matrix is estimated using the data subset with both noisy and clean labels. Then it is used in the model as a dense output layer to correct for the label noise. The network is alternately trained on epochs with the clean dataset with a simple cross-entropy loss and on next epoch with the noisy dataset and a loss corrected with the estimated corruption matrix. The experiments for the prediction of preterm birth at 90 days before delivery showed an improvement in performance compared with baseline and state of-the-art methods.
\end{abstract}
\section{Introduction}
The digitization of hospitals, by the adoption of Electronic Health Record systems, promises to revolutionize the future of healthcare. Several countries have achieved nearly 100\% adoption. Therefore the complexity and size of EHR data is drastically increasing. This is creating new challenges and opportunities to the research community of machine learning for healthcare. In this paper we consider the clinical application of predicting  preterm birth from EHR based on deep learning models. Between 10\% to 15\% of babies are born before 37 weeks of gestation \cite{barros2015distribution}. Preterm birth is the leading cause of mortality and long-term disabilities in neonates.  It is also an important cause of developmental retardation. The cost of preterm deliveries and care exceed 26 billion dollars in the US \cite{behrman2006preterm}. The goal of our application is to predict in advance the risk for a preterm delivery  \cite{tran2016preterm, vovsha2016using}.  Developing such predictive model can be of high value for obstetricians. The availability of large clinical EHR data should help in building accurate and interpretable models.

\section{Related Work}
In supervised learning, it is usually assumed that the data is accurately labeled. However in many real-life applications and especially with massive data, either data or labels may be corrupted by noise. In the case of noisy label, the noise might be class-dependent or instance-dependent. In the former case  each label is flipped  with some fixed probability independently of the instance itself. However in the later case, the label noise depends on the correct hidden label and the instance.

In this work, we are interested in developing deep learning methods that are tolerant to noisy labels. We suppose that the noise affects only the labels, i.e., preterm and full-term births, independently of the EHR mother records. By surveying the literature, we distinguish four approaches in learning with noisy labels  \cite{Frenay2014}: 1) Data Cleansing  by detecting and pre-processing (correcting or removing) the  corrupted labels \cite{Pang2017influence},  2) Loss function reformulation by incorporating the noise in the learning criteria \cite{Natarajan2013, Scott2014, zhang2018generalized}, 3)  Noise robust approaches by using a noise robust loss function such as in SVM, kNNs or logistic regression \cite{Bootkrajang2012,Chen2015} and 4)  Noise tolerant approaches  by using datasets with clean labels to approximate the corruption matrix between the clean and the noisy labels and  then design  a model to predict the correct labels \cite{patrini2017making, Joulin2016, li2017learning, xiao2015learning, han2018co, goldberger2016training, zhang2018generalized, sukhbaatar2014training, ma2018d2l, wang2018iterative}. 

The majority of noise tolerant approaches are based on deep learning models. Sukhbataar et al. \cite{sukhbaatar2014training}  integrated a constrained linear layer at the top of the softmax layer of the base model. Goldberger et al. \cite{goldberger2016training} considered the correct labels as latent variables. Instead of using EM algorithm, they used  a neural network model where the noise is modelled by an additional softmax layer placed between the clean and the noisy output. In the  context of loss correction, Jiang et al. \cite{Jiang2015} and Jindal et al. \cite{Jindal2016}  added regularizers at the end of the network  to  fine-tune the adaptation layer.  Azadi et al. \cite{Azadi2015} proposed  a pre-trained model of AlexNet to extract information  from the  clean data and fine-tune the last layers with a noisy dataset. Yuncheng et al. \cite{Yuncheng2017} developed a distillation framework based on knowledge graph to correct the noisy label. Recently Hendrycks  et al. \cite{hendrycks2018using} proposed a label noise correction method called gold loss correction (GLC).  Given a small trusted dataset $( \mathcal{D}^*)$ and a large untrusted dataset $(\widetilde{\mathcal{D}})$, they first learn  a classifier on the untrusted dataset and determine a label corruption  matrix. Then they train the network on the noisy dataset with a  loss corrected by the corruption matrix and on the clean dataset with a simple cross-entropy loss.  Han et al. \cite{han2018co}  proposed a co-teaching approach by simultaneously  training two networks. At each epoch the weights of one network is updated through the gradient of the other network.
\section{Preterm Dataset with Noisy Labels}
\label{preterm:data}
\begin{wrapfigure}{r}{2.7cm}
\begin{center}
            \begin{tabular}{l|c|c}
             & Pre & Full \\ \hline
            Pre & 0.68 & 0.32 \\ \hline
            Full & 0.2  & 0.8\\ \hline
            \end{tabular}

\end{center}
\makeatletter
\def\@captype{table}
\makeatother
\caption{$\widehat{C}$ estimated using dataset $\mathcal{D}^\prime$.}
\label{fig:Cmat}
\end{wrapfigure}

We used Health Facts$^\copyright$ EMR Data (HF) to extract our pregnancy dataset. The cohort contained mothers with full-term and preterm deliveries from 164 hospitals. The complete diagnosis histories of these subjects were retrieved. Full-term delivery encounters were identified using the following ICD-9 diagnosis codes: (650, 645*, 649.8, 652.5). Preterm deliveries are identified as encounters with one of the following diagnosis codes (644.2*, 640.01). Ambiguous codes related to preterm deliveries were excluded from the criteria. Since the gestational age was not available in the data we used the delivery time as a reference time point. We introduced a prediction period which defines the time gap between the delivery event and the time when the model is asked to compute a prediction score. In this paper, we chose a prediction period of 90 days before delivery. In order to create a realistic prediction scenario and prevent data leakage, all future data with respect to the prediction time point are removed for each subject and made unavailable during training. We excluded subjects with less two visits left for model training. The obtained dataset $\mathcal{D}^*$ has clean labels and gathers in total 23,172 subjects. 

ICD-9 codes on the newborn records are rich in information about the prematurity status. For example in codes 765.2* and  765.1*, the fifth-digit sub-classifications indicates respectively the gestational age at delivery and the baby weight. Since HF is a de-identified EHR database, it is not possible to link mothers to their babies records and thus we cannot augment mother records with additional information on the deliveries. We propose a simple and fast algorithm to re-link babies to mothers based on admission and discharge times. We note that the timestamps in HF has not been manipulated during de-identification. For each hospital, mother deliveries and newborn events are identified based on ICD codes. We have extracted 739,000 deliveries and 221,000 newborns. We have restricted the selection of the newborns to the ones that can be categorized into full-term or preterm newborns. We defined the time vector $t=[t_{adm},\ t_{dis}]$, where $t_{adm}$ and $t_{dis}$ are respectively the admission and discharge times. We used $t$ to find, for each hospital, the nearest mother in time to each newborn baby. As one mother could be the nearest to multiple babies, we applied a threshold of 3 (in case of multiple gestations) and an $L_1$ time distance of 24 hours to exclude the unlikely candidate mothers. The time threshold is justified by the fact that there might delays in the EHR system in capturing  mothers and babies admission and discharge times. After applying the filtering criteria, the matching algorithm resulted in 23,578 mother-baby links allowing to extend the labeling of the mother deliveries as preterm or full-term. This constitutes our dataset $\widetilde{\mathcal{D}}$ with noisy labels. The obtained labels are considered noisy because the linkage algorithm is based solely on temporal information and hence can lead to erroneous matching.

\section{Alternating Loss Correction}
\label{gen_inst}

As explained in Section \ref{preterm:data}, we have two datasets available for the prediction of preterm births: a dataset $\mathcal{D}^*$ with clean labels and a dataset $\widetilde{\mathcal{D}}$ with noisy labels. The label corruption is specified by a distribution $p\left(\tilde{y}\mid y^*\right)$ independent of the input $x$. Since we have a binary classification problem, we can estimate the $2\times 2$ corruption matrix defined by $C_{ij}=p\left(\tilde{y}=j,\ y^*=i\right)$ using the subset $\mathcal{D}^\prime$ with both clean and noisy labels.  $\mathcal{D}^\prime$ has 2,133 subjects.  We used this subset to estimate the matrix $\widehat{C}$ of $C$ using equation (\ref{eq:C}).
Pre-training methods for noisy labels suffer from memorizing data with noisy labels. On the other hands Pre-training with clean data and fine tuning with noisy data tend to have poor performance. In order to circumvent this issue, we propose to alternate the use of both datasets during training. A loss correction is applied while training with $\widetilde{\mathcal{D}}$. A simple cross entropy loss is used when the model is trainined with $\mathcal{D}^*$.

\begin{minipage}{6cm}
 \hspace{0pt}

\begin{equation}
\widehat{C}_{ij}=\frac{1}{|A_i|}{\sum_{x \in A_i} \mathbbm{1}{\left\{\tilde{y} = j\mid x\right\}}}
\label{eq:C}
\end{equation}
 where  $A_i$ is the subset of $\mathcal{D}^\prime$ with clean label $i$. The classifier $f$ is a deep neural network similar to RETAIN \cite{choi2016retain}. The output layer of $f$ is a dense layer with a softmax activation having $2\times2$ parameters. The loss correction can be viewed as a dense layer of constant weights $\widehat{C}_{ij}$ and zero bias. The network $f$ is alternately trained on an epoch using $\mathcal{D}^*$ using a simple cross-entropy loss and on the next epoch using  $\widetilde{\mathcal{D}}$ with a loss correction.
\end{minipage}
\hfill%
\begin{minipage}{7.5cm}
  \vspace{0pt}  
\begin{algorithm}[H]
\SetKwInOut{Input}{Input}
\SetKwInOut{Output}{Output}
\Input{$\mathcal{D}^*:\left(x, y^*\right)$ dataset of clean labels;
$\widetilde{\mathcal{D}}: \left(x, \Tilde{y}\right)$ dataset of noisy labels; \\
$\mathcal{D}^\prime:\left(x, y^\prime\right)$ dataset with both noisy and \\\hspace{1.65cm}clean labels; \\
\\ $N_e$: number of training epochs.}
{  Estimate $\widehat{C}$ with (\ref{eq:C}) using dataset $\mathcal{D}^\prime$ \\
Initialize Network $f(x) = \widehat{p}\left(y\mid x; \theta_0 \right)$ \\
 \For{$epoch = 0, \ldots , N_e-1 \ $}{
    \uIf{epoch is odd}{
    Train  $f$ with $\mathcal{D}^*$ with loss $\ell(f(x), y^*))$\;
  }
  \Else{
 Train $f$ with $\widetilde{\mathcal{D}}$ with loss $\ell(\widehat{C} f(x), \tilde{y}))$\;
  }
  }
  }
\Output{ Model $f(x)=\widehat{p}\left(y\mid x;\widehat{\theta}\right)$}
\caption{{Alternating Loss Correction (ALC)}}
\end{algorithm}
\end{minipage}

\section{Experiments}
\label{headings}
\begin{wrapfigure}{r}{7cm}
\makeatletter
\def\@captype{table}
\makeatother
\vspace{-1cm}
\caption{Summary results in terms of Area Under ROC and Area Under Precision-Recall Curve.}
\label{tab:result}
\centering
\begin{tabular}{lll}
 \toprule
Method    & AUC     & PR-UC \\
\midrule
ALC - $\mathcal{D}^*\sim \widetilde{\mathcal{D}}$ & {\bf 82.79$\pm$0.72}&  {\bf 73.05$\pm$1.37} \\
GLC - $\widetilde{\mathcal{D}}$ then $\mathcal{D}^*$ & 80.77$\pm$1.44&  71.07$\pm$2.05\\
GLC - $\mathcal{D}^*$ then $\widetilde{\mathcal{D}}$ & 64.13$\pm$1.59&  49.48$\pm$3.03\\
No-LC - $\mathcal{D}^*+\widetilde{\mathcal{D}}$ & 77.87$\pm$0.87& 66.65$\pm$1.76     \\
No-LC - $\mathcal{D}^*$ & 80.71$\pm$1.49&  72.11$\pm$2.04\\
No-LC - $\widetilde{\mathcal{D}}$ & 71.20$\pm$1.41&  59.72$\pm$1.68\\
\bottomrule
\end{tabular}
\end{wrapfigure}
 The sequence of diagnosis codes are ordered in the patient-visit timeline. Patient encounters of the same day are merged to reduce the temporal frequency in the data. The discrete ICD codes, represented by 17629-dimensional hot vector, are then passed into a 200-dimensional floating-point embedding. For each batch, the sequences are padded with zeros to have the same number of visits.
 The embedding vectors of the different codes within the same visit are summed up. Then they are passed to a Recurrent Neural Network with an attention mechanism.  The network architecture is a modified version of RETAIN model \cite{choi2016retain}.  We benchmarked the proposed ALC algorithm with 1) Gold Loss Correction (GLC) method \cite{hendrycks2018using} and 2) training without loss correction. All evaluations reported in this work are based on validation and test sets from the dataset with clean labels $\mathcal{D}^*$.   We considered few training scenarios in our benchmark: GLC is trained first with the noisy label dataset $\widetilde{\mathcal{D}}$ using loss correction then the model further trained with the clean-labels dataset $\mathcal{D}^*$. This configuration is the one implemented in GLC. We evaluated GLC by flipping the order of the datasets, i.e., we first trained with $\mathcal{D}^*$ without loss correction and further trained the model on $\widetilde{\mathcal{D}}$ with loss correction. We evaluated also the baseline scenario where only 1) clean labels are used and samples of clean and noisy labels are mixed. In both cases, no loss correction is applied. This is denoted by No-LC in table \ref{tab:result}. The models have been first evaluated with few different number of epochs $N_e$. Training with 10 epochs was enough to obtain result stabilization. The models were implemented using Keras 2.1.6 with Tensorflow 1.8.0 back-end. We trained the model in parallel using two Tesla V100 GPUs. It took approximately 2-3 mins to train a model on 10 epochs.  Table \ref{tab:result} summarizes the results in terms of Area under ROC (AUC) and Area under Precision Recall curve (PR-UC). The splitting of data into training, validation and test sets was randomly repeated 20 times to obtain performance means and standard deviations. The proposed ALC achieved the best results in the benchmark. By alternating through both datasets, ALC was able to leverage on sample from noisy labels and achieve a gain in performance compared with the baseline using only clean labels (No-LC - $\mathcal{D}^*$). ALC is followed by GLC - $\widetilde{\mathcal{D}}$ then $\mathcal{D}^*$ when it is first trained on noisy labels with corrected loss then on clean labels without correction. GLC - $\mathcal{D}^*$ then $\widetilde{\mathcal{D}}$ gave the worst result and seems to under-fit the data. This could be explained by the fact that the model tends to memorize the latest examples seen during training. The baseline results, No-LC - $\mathcal{D}^*$, solely on clean labels is 80.71 (AUC) and 72.11 (PR-UC). The data augmentation (No-LC - $\mathcal{D}^*+\widetilde{\mathcal{D}}$) by merging both datasets  $\mathcal{D}^*$ and $\widetilde{\mathcal{D}}$ resulted in a performance degradation.
\label{others}


\section{Discussion and Future Work}
The problem of matching of mothers and babies is relevant in general for massive de-identified EHR datasets and can be useful for studying pregnancy outcomes beyond the scope our application of predicting preterm births. Our simple matching heuristic  has achieved an accuracy of 72\% on the labels measured on $\mathcal{D}^\prime$. This could be further improved by considering a more global matching method based on linear optimization. The ambiguous ICD-9 codes that were excluded from the cohort definition could be used to add subjects with noisy labels. In a future work, we will consider the use of probabilistic noise model where the label values continuously range between 0 and 1 for binary classification. The values will reflect the noise level on the labels. Values of 0 or 1 indicate the clean labels and the rest are noisy with different levels of noise. A label of 0.5 has the highest uncertainty on the label assignment. This can help increasing the size of the training set.  In this context, we plan to combine a regression problem on the continuous noisy labels and a classification problem on the clean labels. We mention that in the case where $\mathcal{D}^\prime$ is very small, we can use a similar approach proposed in \cite{hendrycks2018using} to estimate the corruption matrix $C$.
\section{Conclusion}
In this paper, we were interested by the prediction of preterm birth using diagnosis information from de-identifed EHR data. We have devised a heuristic to match mothers and babies and hence augment our cohort with additional data examples with noisy labels. Then we have  introduced a new learning algorithm called Alternating Loss Correction (ALC) to robustly  train  deep neural networks with noisy labels. The idea behind ALC is to involve both clean and noisy labels in an alternating fashion during training to avoid over-fitting and increase the generalization capability of the model.  In ALC algorithm, we first estimate the corruption matrix  on the data subset that have both clean and noisy labels, then we train  the model on the clean and noisy datasets by alternating the losses. ALC achieved an improvement in prediction performance compared with baseline and state-of-the-art methods. It could be generalized to the training of deep learning models with multiple datasets.

\bibliography{bib_ml4h.bib}

\begin{thebibliography}{10}

\bibitem{barros2015distribution}
Fernando~C Barros, Aris~T Papageorghiou, Cesar~G Victora, Julia~A Noble, Ruyan
  Pang, Jay Iams, Leila~Cheikh Ismail, Robert~L Goldenberg, Ann Lambert,
  Michael~S Kramer, et~al.
\newblock The distribution of clinical phenotypes of preterm birth syndrome:
  implications for prevention.
\newblock {\em JAMA pediatrics}, 169(3):220--229, 2015.

\bibitem{behrman2006preterm}
Richard Behrman and A~Butler.
\newblock Preterm birth: Causes consequences and prevention. committee on
  understanding premature birth and assuring health outcomes, institute of
  medicine of the national academies, 2006.

\bibitem{tran2016preterm}
Truyen Tran, Wei Luo, Dinh Phung, Jonathan Morris, Kristen Rickard, and Svetha
  Venkatesh.
\newblock Preterm birth prediction: Deriving stable and interpretable rules
  from high dimensional data.
\newblock In {\em Conference on machine learning in healthcare, LA, USA}, 2016.

\bibitem{vovsha2016using}
Ilia Vovsha, Ansaf Salleb-Aouissi, Anita Raja, Thomas Koch, Alex Rybchuk,
  Axinia Radeva, Ashwath Rajan, Yiwen Huang, Hatim Diab, Ashish Tomar, et~al.
\newblock Using kernel methods and model selection for prediction of preterm
  birth.
\newblock {\em arXiv preprint arXiv:1607.07959}, 2016.

\bibitem{Frenay2014}
B.~Frenay and M.~Verleysen.
\newblock Classification in the presence of label noise: a survey.
\newblock 25(5):845–869, 2015.

\bibitem{Pang2017influence}
Pang~Wei Koh and Percy Liang.
\newblock Understanding black-box predictions via influence functions.
\newblock In {\em Proc. of the 34th International Conference on Machine
  Learning, volume 70 of Proceedings of Machine Learning Research}, page
  1885–1894, 2017.

\bibitem{Natarajan2013}
Pradeep K~Ravikumar Nagarajan~Natarajan, Inderjit S~Dhillon and Ambuj Tewari.
\newblock Learning with noisylabels.
\newblock In {\em In C. J. C. Burges, L. Bottou, M. Welling, Z. Ghahramani, and
  K. Q. Weinberger, editors, Advances in Neural Information Processing Systems
  26, pages. Curran Associates, Inc., 2013}, page 1196–1204, 2013.

\bibitem{Scott2014}
Dragomir Anguelov Christian Szegedy Dumitru~Erhan Scott E.~Reed, Honglak~Lee
  and Andrew Rabinovich.
\newblock Training deep neural networks on noisy labels with bootstrapping.
\newblock In {\em In C. J. C. Burges, L. Bottou, M. Welling, Z. Ghahramani, and
  K. Q. Weinberger, editors, Advances in Neural Information Processing Systems
  26, pages. Curran Associates, Inc., 2013}, page 1196–1204, 2015.

\bibitem{zhang2018generalized}
Zhilu Zhang and Mert~R Sabuncu.
\newblock Generalized cross entropy loss for training deep neural networks with
  noisy labels.
\newblock {\em arXiv preprint arXiv:1805.07836}, 2018.

\bibitem{Bootkrajang2012}
Jakramate Bootkrajang and Ata Kaban.
\newblock Label-noise robust logistic regression and its applications.
\newblock In {\em In Joint European Conference on Machine Learning and
  Knowledge Discovery in Databases}, page 143–158. Springer, 2012.

\bibitem{Chen2015}
X.~Chen and A.~Gupta. s.
\newblock Webly supervised learning of convolutional network.
\newblock In {\em Proc. Proceedings of the IEEE International Conference on
  Computer Vision}, pages 1431--1439, 2015.

\bibitem{patrini2017making}
Giorgio Patrini, Alessandro Rozza, Aditya~Krishna Menon, Richard Nock, and
  Lizhen Qu.
\newblock Making deep neural networks robust to label noise: A loss correction
  approach.
\newblock In {\em Proc. IEEE Conf. Comput. Vis. Pattern Recognit.(CVPR)}, pages
  2233--2241, 2017.

\bibitem{Joulin2016}
A.~Jabri A.~Joulin, L. van der~Maaten and N.~Vasilache.
\newblock Learning visual features from large weakly supervised data.
\newblock In {\em In European Conference on Computer Vision}, page 67–84.
  Springer, 2016.

\bibitem{li2017learning}
Yuncheng Li, Jianchao Yang, Yale Song, Liangliang Cao, Jiebo Luo, and Li-Jia
  Li.
\newblock Learning from noisy labels with distillation.
\newblock In {\em ICCV}, pages 1928--1936, 2017.

\bibitem{xiao2015learning}
Tong Xiao, Tian Xia, Yi~Yang, Chang Huang, and Xiaogang Wang.
\newblock Learning from massive noisy labeled data for image classification.
\newblock In {\em Proceedings of the IEEE Conference on Computer Vision and
  Pattern Recognition}, pages 2691--2699, 2015.

\bibitem{han2018co}
Bo~Han, Quanming Yao, Xingrui Yu, Gang Niu, Miao Xu, Weihua Hu, Ivor Tsang, and
  Masashi Sugiyama.
\newblock Co-teaching: Robust training of deep neural networks with extremely
  noisy labels.
\newblock {\em Advances in Neural Information Processing Systems}, 2018.

\bibitem{goldberger2016training}
Jacob Goldberger and Ehud Ben-Reuven.
\newblock Training deep neural-networks using a noise adaptation layer.
\newblock {\em International Conference on Learning Representation}, 2016.

\bibitem{sukhbaatar2014training}
Sainbayar Sukhbaatar, Joan Bruna, Manohar Paluri, Lubomir Bourdev, and Rob
  Fergus.
\newblock Training convolutional networks with noisy labels.
\newblock {\em arXiv preprint arXiv:1406.2080}, 2014.

\bibitem{ma2018d2l}
Xingjun Ma, Yisen Wang, Michael~E Houle, Shuo Zhou, Sarah~M Erfani, Shu-Tao
  Xia, Sudanthi Wijewickrema, and James Bailey.
\newblock Dimensionality-driven learning with noisy labels.
\newblock {\em ICML}, 2018.

\bibitem{wang2018iterative}
Yisen Wang, Weiyang Liu, Xingjun Ma, James Bailey, Hongyuan Zha, Le~Song, and
  Shu-Tao Xia.
\newblock Iterative learning with open-set noisy labels.
\newblock In {\em CVPR}, 2018.

\bibitem{Jiang2015}
Thomas Leung Li-Jia~Li Lu~Jiang, Zhengyuan~Zhou and Li~Fei-Fei.
\newblock Mentornet: Regularizing verdeep neural networks on corrupted labels.
\newblock In {\em arXiv preprint arXiv:1712.05055}, 2017.

\bibitem{Jindal2016}
Matthew~Nokleby Ishan~Jindal and Xuewen Chen.
\newblock Learning deep networks from noisy labels with dropout regularization.
\newblock In {\em Data Mining (ICDM), 2016 IEEE 16th International Conference},
  page 967–972, 2016.

\bibitem{Azadi2015}
S.~Jegelka S.~Azadi, J.~Feng and T.~Darrell.
\newblock Auxiliary image regularization for deep cnns with noisy labels.
\newblock In {\em arXiv preprint arXiv:1511.07069}, 2015.

\bibitem{Yuncheng2017}
Yale Song Liangliang Cao Jiebo~Luo Yuncheng~Li, Jianchao~Yang and Jia Li.
\newblock Learning from noisy labels with distillation.
\newblock In {\em arXiv preprint arXiv:1703.02391}, 2017.

\bibitem{hendrycks2018using}
Dan Hendrycks, Mantas Mazeika, Duncan Wilson, and Kevin Gimpel.
\newblock Using trusted data to train deep networks on labels corrupted by
  severe noise.
\newblock 2018.

\bibitem{choi2016retain}
Edward Choi, Mohammad~Taha Bahadori, Jimeng Sun, Joshua Kulas, Andy Schuetz,
  and Walter Stewart.
\newblock Retain: An interpretable predictive model for healthcare using
  reverse time attention mechanism.
\newblock In {\em Advances in Neural Information Processing Systems}, pages
  3504--3512, 2016.

\end{thebibliography}

\end{document}